\documentclass[review]{elsarticle}

\usepackage{lineno,hyperref}

\usepackage{amsmath}
\usepackage{amsfonts}
\usepackage[noend]{algpseudocode}
\usepackage{float}
\usepackage{hyperref}
\usepackage{algorithm}
\usepackage{booktabs}
\modulolinenumbers[5]

\begin{document}
\begin{frontmatter}

\title{Subsampled Turbulence Removal Network}

\author[fn1]{Wai Ho Chak}
\ead{whchak@math.cuhk.edu.hk}
\author[fn1]{Chun Pong Lau }
\ead{cplau@math.cuhk.edu.hk}
\author[fn1]{Lok Ming Lui, \corref{cor}}
\ead{lmlui@math.cuhk.edu.hk}
\address[fn1]{Department of Mathematics, The Chinese University of Hong Kong, Hong Kong}
\cortext[cor]{Corresponding author}
 
\begin{abstract}
We present a deep-learning approach to restore a sequence of turbulence-distorted images from turbulent deformations and space-time varying blurs. Instead of requiring a massive training sample size in deep networks, we propose a training strategy that is based on a data augmentation method to model the turbulence from a relatively small dataset. Then we incorporate a subsampling method into the deep network to enhance the restoration performance of the trained {\bf GAN} model. The contribution of the paper is threefold: First, we introduce a simple but effective data augmentation algorithm to model the turbulence for training in the deep network; Second, we propose the {\bf W}asserstein {\bf GAN} combined with the multiframe input and $\ell_1$ cost for successful restoration of turbulence-corrupted video sequence; Third, we incorporate a subsampling algorithm into the deep network to filter out strongly corrupted frames so as to obtain an improved restored image. 
\end{abstract}
\end{frontmatter}

\section{Introduction}
The problem of image restoration from a sequence of frames under the atmospheric turbulence is challenging due to the dramatic downgrade in the image quality from the geometric distortions and the space-time varying blurs. Multiple factors such as temperature changes, air turbulent flow, densities of air particles, carbon dioxide level and humidity lead to the occurrence of several turbulence layers with various changes in the refractive index \cite{hufnagel1964modulation} \cite{roggemann2018imaging}. These factors together explain the higher chance of obtaining corrupted video sequences in locations where the variation among these factors is large. In practice, either techniques in hardware-based adaptive optics \cite{pearson1976atmospheric} \cite{tyson2010principles} or methods in image processing \cite{shimizu2008super} \cite{seitz2009filter} \cite{li2007atmospheric} \cite{vorontsov1999parallel} \cite{hirsch2010efficient} are employed to remove the turbulence distortion in the images, but those prevailing models from either way can barely address to the majority of these factors. \par
Due to the fact that the atmospheric turbulence is complicated to be modeled, a deep learning approach which does not heavily require the underlying assumptions is more reasonable to tackle the problem than models relying on certain assumptions on the turbulence. We are thus motivated to investigate the possibility to remove geometric distortions and restore a good-quality image by using a generative model that does not explicitly take the above-mentioned factors into consideration. However, the unavailability of massive turbulence-distorted video frames disables the application of deep learning approaches to tackle the problem.  \par
In this paper, we introduce a simple and yet effective data augmentation method to overcome the problem of data scarcity. The method models real turbulence with different deformations and different extent of blurs in order to provide sufficient training data. Since the artificial turbulence is randomly generated with different strength of deformations and blurs, a variety of turbulence-distorted videos can be produced from a single image. In general, it is known that the performance of image restoration is commensurate with the training sample size. Nevertheless, with the data augmentation method, the size requirement of the training data is not too restrictive and demanding in our proposed deep network to restore the turbulence-distorted images. \par
With the augmented training data, a deep network can be trained to solve the deturbulence problem. We propose a subsampled Wasserstein Generative Adversarial Network ({\bf WGAN}) with multiframe input and $\ell_1$ cost to simultaneously remove geometric distortions and blurring effects of turbulence-distorted image sequences. {\bf WGAN} is known for its effectiveness in generating a clear image from noises. Together with the $\ell_1$ cost applied to the network, important features of the images can be restored even though they are corrupted. To gather enough information, it is natural to take multiple frames from the video as the input of the turbulence-removal network. Using multiple frames as input is essential to obtain a clear image from a turbulence-distorted video. \par
In the testing stage, we propose to incorporate a subsampling algorithm to the trained network for better performance. Usually, turbulence-distorted video consists of mildly distorted frames. The subsampling method extracts those sharp and mildly distorted frames in order to achieve an even better restoration result. We experimentally show that by incorporating the subsampling method, the performance of removing geometric distortions and blurs of the degraded images can be significantly improved. 

\subsection{Contributions}
The main contributions of this paper are listed as follows:
\begin{enumerate}
\item[1]
We propose a deep-learning approach, which is a {\bf WGAN} model with the multiframe input and $\ell_1$ loss, for the restoration of turbulence-distorted images. To the best of our knowledge, it is the first work to study the feasibility of applying deep convolutional neural network for solving the deturbulence problem to simultaneously remove geometric distortions and space-time varying blurs.
\item[2] We propose a data augmentation method to generate geometrically distorted and blurry images for training. It overcomes the problem of data scarcity. As such, the use of deep learning approaches to tackle the deturbulence problem is made possible that a sufficiently large dataset is not required.
\item[3] We propose to incorporate a subsampling method into the trained network to obtain a better restored image. Experimental results demonstrate that the performance of the proposed model can be significantly improved with the subsampling strategy.  
\end{enumerate}

\section{Related Work}

\subsection{Restoration of turbulence-distorted images}
The main tasks of restoring turbulence-corrupted images consist of the removal of geometric distortions and space-time varying blurs. It is in general challenging to discard both the geometric distortions and blurs simultaneously. Several previous works are firstly devoted to reconstruct a clean image through the process of image fusion by registering the image frames to a good reference image. Meinhardt-Llopis and Micheli \cite{meinhardt2014implementation}  \cite{micheli2014linear} proposed a reference extraction method by registering frames to a `centroid' image. The basic idea is to warp each image frame by the average deformation field between it and the other images from the turbulence-degraded video. This method has an assumption that the deformation between the original image and the distorted frames is zero on average. However, the estimated movements of individual pixels can sometimes be much larger, and the mean displacement of each pixel may deviate more significantly from zero in a real turbulence-distorted video. It may pose a challenge for the centroid method to remove all the geometric distortions. Another approach to obtain a clear reference image is done by selecting a "lucky frame", which is the sharpest frame from a distorted video \cite{vorontsov2001anisoplanatic}. This method is motivated and supported by the statistical proofs \cite{fried1978probability} that show a high probability of extracting video frames with sharp texture details given a sufficient amount of frames. Nevertheless, in many situations, getting a frame that is entirely sharp everywhere is difficult. To alleviate this issue, the Lucky-Region method has been proposed by Aubailly {\it et al.} \cite{aubailly2009automated}, which chooses the sharpest patch from each frame and combine them afterward. Motivated by this patch-wise sharpness selection method, another approach introduced by Anantrasirichai {\it et al.} \cite{anantrasirichai2013atmospheric} suggests to having a frame selection prior to registration. A composite cost function was introduced, and the selection was done in one step by sorting. However, some of the selected frame may geometrically differ significantly from the reference image. On the other hand, the cost function assumes the reference image given by the temporal intensity mean over all frames can accurately approximate the underlying true image, which is usually not the case. Similarly, a subsampling method introduced by Roggemann \cite{roggemann1994image} selects subsamples from images produced by adaptive-optics systems to generate a temporal mean with higher signal-to-noise ratio.

To enhance the accuracy of the registration onto a reference image, a feasible approach is to stabilize the video and reduce the deformation between each frame and the reference image.  The SGL method purposed by Lou {\it et al.} \cite{lou2013video} incorporates Sobolev gradient and Laplacian for stabilization of the video sequence, and finds the latent image by the Lucky-Region method.

Robust Principle Component Analysis (RPCA) \cite{candes2011robust} is a recent approach to solve the deturbulence problem. Low-rank decomposition method proposed by He {\it et al.} \cite{he2016atmospheric} decomposes the video sequence into the low-rank and sparse parts. A variational approach introduced by Xie {\it et al.} \cite{xie2016removing} is applied to improve the initial reference image as the low-rank image that captures the texture information and suppresses geometric distortions, although it usually looks blurry. Registration may sometimes fail when there is a large deformation between the observed video frames and the reference image.\par

Another recent approach is the joint subsampling and reconstruction variational model proposed by Lau {\it et al.} \cite{lau2017variational}. An advantage of the model is that there is no registration involved during the subsampling and reconstruction processes, and hence it is computationally efficient. Using the proposed energy model with various fidelity terms, restoration of turbulence-distorted images of different degrees of distortions can be achieved.

\subsection{Generative Adversarial Networks}
Generative adversarial networks ({\bf GAN}s) firstly proposed by Goodfellow at al. \cite{goodfellow2014generative} defines two separated competitors: the generator $G_{\theta}$ and the discriminator $D_{\xi} $. The generator is designed to produce samples from noise $\mathcal{Z}$ while a discriminator is designed to distinguish real sample $y_i$ and generated sample $G_\theta(z_i)$. The main objective of the generator is to generate perceptually persuasive samples that are challenging to be discriminated by the real samples. The competition between the generator $G$ and the discriminator $D$ can be described by the minimax objective shown as follows:
\begin{equation}
\underset{G}{\min}  \underset{D}{\max} \quad \underset{x  \sim \mathbb{P}_r} { \mathbb{E} } [\log D(x)] + \underset{\tilde{x}  \sim \mathbb{P}_g} { \mathbb{E} } [\log (1-D(\tilde{x}))] 
\end{equation}
where $ \mathbb{P}_r $ is the data distribution and $\mathbb{P}_g$ is the generated distribution given by $\tilde{x}  = G(z), z \sim P(z)$, where $z$ is sampled from a noise distribution. The advantage of {\bf GAN}s is the ability to generate clear samples with high perceptual quality. However, as described by Salimans {\it et al.} \cite{salimans2016improved}, there are undesirable issues such as vanishing gradients and mode collapse in the training. The difficulties can be explained by the fact that minimizing the objective function for {\bf GAN}s is equivalent to minimizing the Jensen-Shannon divergence, which is locally saturated and results in vanishing gradients, between the data and model distributions. \par
Later, Arjovsky {\it et al.} \cite{arjovsky2017wasserstein} addressed the gradient vanishing problem by introducing the weaker Wasserstain-1 distance $W(\mathbb{P}_r, \mathbb{P}_g)$ which gives clear gradients almost everywhere in the {\bf GAN} model. The competition between the two networks is reformulated as our minimax optimization objective:
\begin{equation}
\underset{G}{\min}  \underset{D \in \mathcal{D}}{\max}  \quad \underset{x  \sim \mathbb{P}_r} { \mathbb{E} } [ D(x)] - \underset{\tilde{x}  \sim \mathbb{P}_g} { \mathbb{E} } [ D(\tilde{x})] 
\end{equation}
where $\mathcal{D}$ is the set of $1$-Lipschitz functions such that ${\Vert D \Vert}_L \leq 1$. The original Lipschitz constraint enforcement proposed by Arjovsky at al.  \cite{arjovsky2017wasserstein} is weight clipping to $[-c,c]$. Another approach proposed by Gulrajani {\it et al. } \cite{gulrajani2017improved} is adding the gradient penality term
\begin{equation}
\lambda \underset{\tilde{x}  \sim \mathbb{P}_{\tilde{x}}} { \mathbb{E} } [ {( {\Vert \nabla_{\tilde{x}} D(\tilde{x})\Vert}_2-1)}^2 ]
\end{equation}
The approach does not require hyperparameter tuning and is robust to the selection of the architecture of generator. In contrast to the conventional convolutional neural network, {\bf GAN}s can generate clearer images. {\bf WGAN}-$\ell_1$ proposed by Kupyn {\it et al.} \cite{kupyn2017deblurgan} has been shown effective in image deblurring.

\section{{\bf TRN}: Turbulence Removal Network}
\begin{figure}[!htbp]
\centering
\includegraphics[scale=0.38]{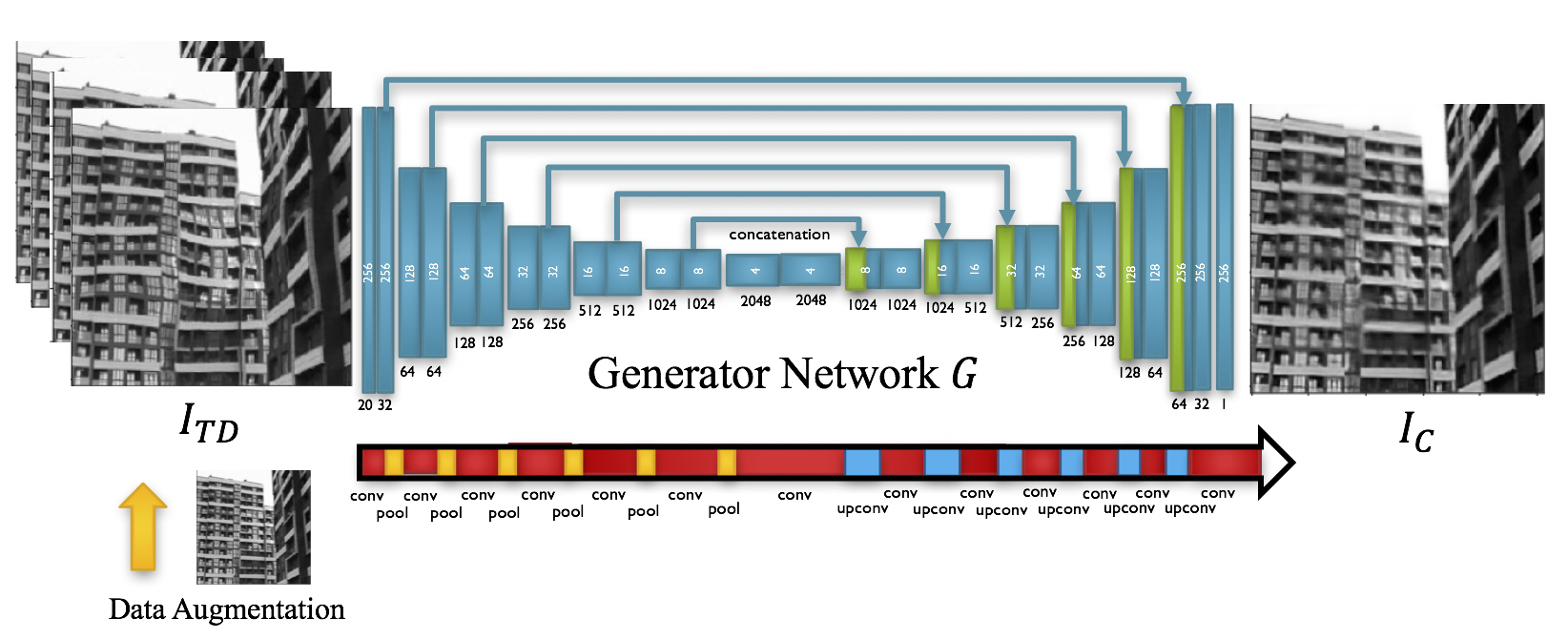}
\centering
\includegraphics[scale=0.38]{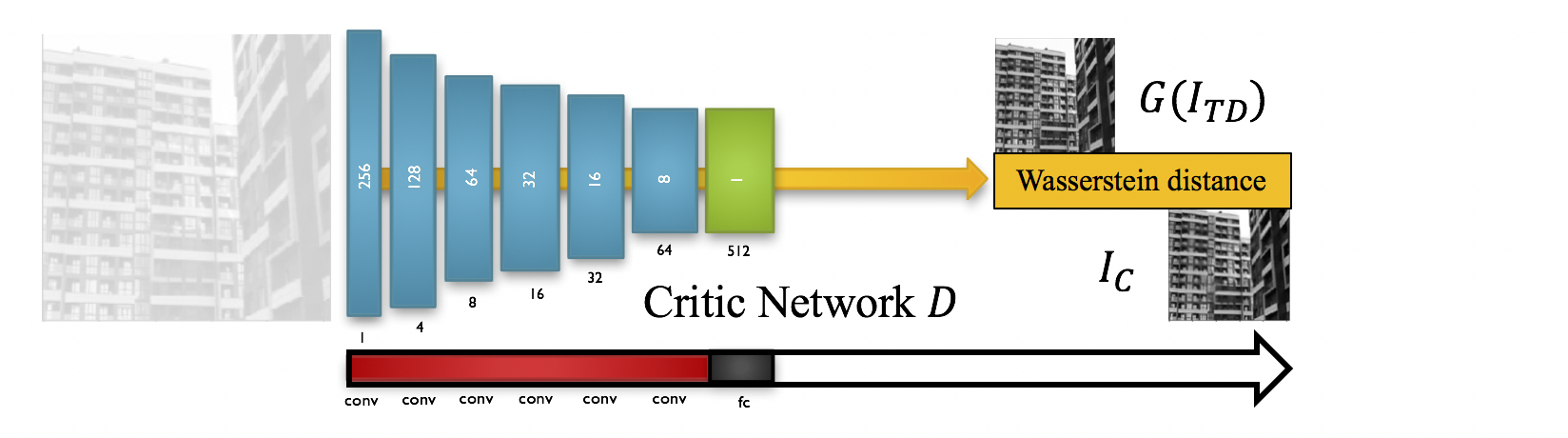}
\caption{The generator network $G$ has the U-net architecture, and the critic network $D$ is the conventional convolutional neural network. The subsampled frames are concatenated before passing through the generator network.}
\label{fig:net}   
\end{figure}

In this section, we describe our proposed method based on deep convolutional neural network, namely, the {\bf T}urbulence {\bf R}emoval {\bf N}etwork ({\bf TRN}). Figure \ref{fig:net} shows that whole network architecture for both the generator network $G$ and the critic network $D$. 

\subsection{Data Augmentation for Turbulence}
The first step of our proposed algorithm is to synthesize sufficient training data distorted by turbulence. A large sample size of training data is typically necessary for solving tasks using deep learning approaches, but unfortunately there is limited turbulence-distorted data available. This hinders the application of deep learning approaches for turbulence removal. To alleviate this issue, we introduce a new, simple but effective method for generating the training data from a few data for deep learning.\par
More specifically, a single (clean) frame $I$ is transformed to another image $I^t$ with geometric distortions and blurs as follows. We first select $Q = (w-2N) \times (w-2N)$ pixel positions randomly, where $w$ and $h$ are the width and height of the image respectively. At each randomly selected pixel position $(x,y)\in \mathcal{R}$, we consider a $N\times N$ local patch $P_{x,y}^N$ around the pixel. A motion vector field $V_{x,y} = (u, v)$ is then generated in $P_{x,y}^N$. For each $p\in P_{x,y}^N$, the vector $(u(p), v(p))$ is sampled from a normal distribution, smoothened by a Gaussian kernel and entry-wisely multiplied by a strength value of distortion. Mathematically, the vector field $V_{x,y}$ can be written as:
\begin{equation}
    V_{x,y} = S\ (G_{\sigma}*\mathcal{N}_1,G_{\sigma}*\mathcal{N}_2),
\end{equation}
\noindent where $G_{\sigma}$ is the Gaussian kernel with standard deviation $\sigma$, $S$ is the strength value, $\mathcal{N}_1$ and $\mathcal{N}_2$ are randomly selected from a normal distribution. $V_{x,y}$ is then extended to the whole image domain by setting zero outside $P_{x,y}^N$. It is then employed to wrap the original image $I$ to get a transformed image. We repeat this process by $M$ iterations.

Essentially, the overall motion vector field $V=(u,v)$ after $M$ iterations is defined by fusing the vector patches together wherever overlapping. Mathematically,
\begin{equation}
    V = \sum_{(x,y)\in \mathcal{R}} V_{x,y},
\end{equation}
\noindent where $\mathcal{R}$ is the collection of $Q$ randomly selected pixel positions.

\begin{figure*}[!htbp]
\centering
\begin{tabular}{cccc}
 \includegraphics[scale=0.38]{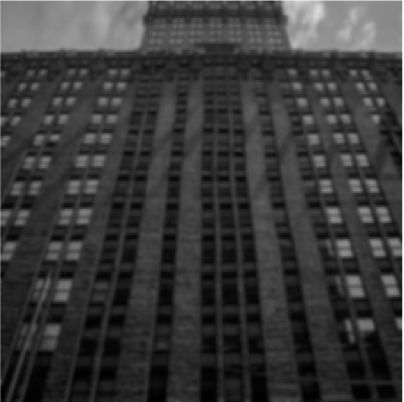}&
\includegraphics[scale=0.38]{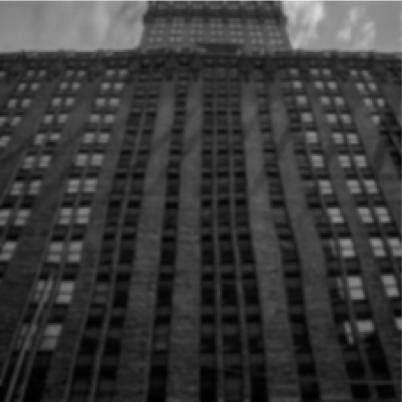}&
\includegraphics[scale=0.38]{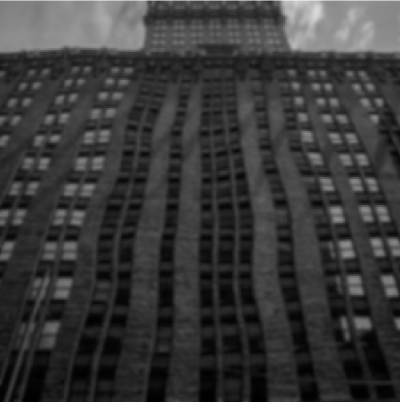}&
\includegraphics[scale=0.38]{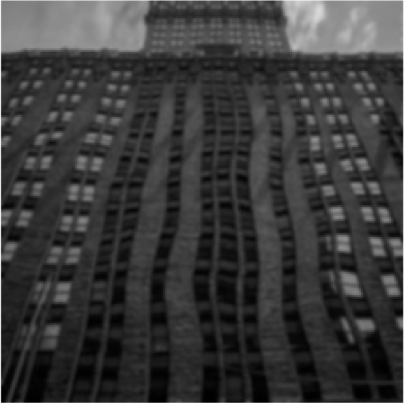}\\
   (a) \small{S = 0.1} & (b) \small{S = 0.2} & (c) \small{S = 0.3} & (d) \small{S = 0.4} \\
\end{tabular} 
\caption{Generation of turbulence-distorted frames with different distortion strength $S$ and fixed blur constant $B = 1$.}
\label{fig:distortion_strength} 
\end{figure*}

We denote the transformed image by $I^t$. The transformed image $I^t$ is further blurred by a Gaussian kernel $w(n) = \exp(-\frac{n^2}{2 B^2})$. The parameter $B$ is sampled uniformly from $[0.1,1]$. The final transformed image with geometric distortions and blurs is given by $I^i = w*I^t$. Using the proposed algorithm with randomized parameters, an original clean image $I^C$ is transformed into a sequence of images $\{I_1,...,I_n\}$ with geometric distortions and blurs for training. Figure \ref{fig:distortion_strength} shows some of the transformed images with different strength values. Experimental results suggest that this proposed algorithm with random parameters can successfully cover most possible deformations and hence geometric distortions can be successfully learnt from the deep network. 

\begin{figure*}[!htbp]
\centering
\includegraphics[scale=0.45]{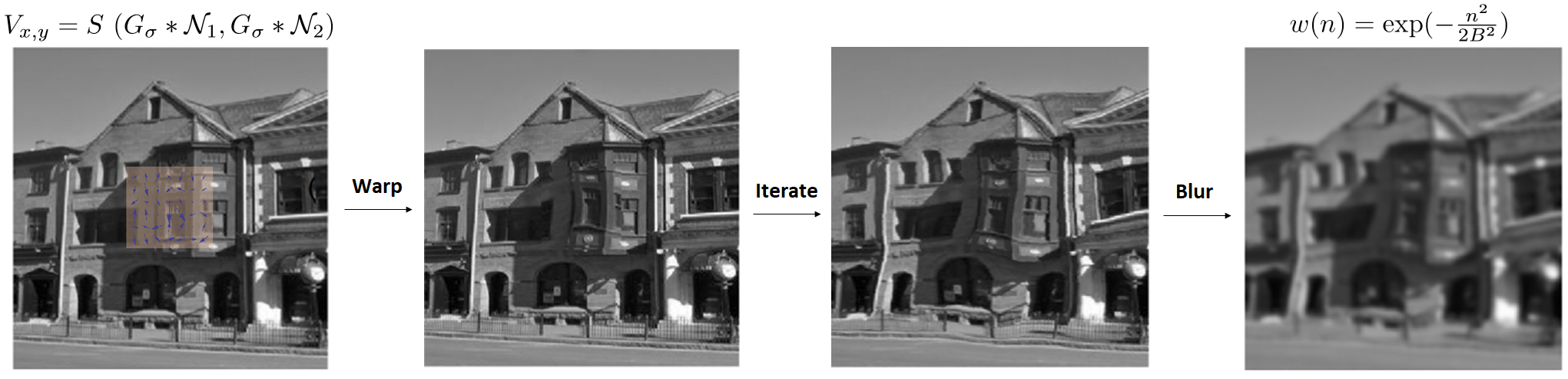}
\caption{The overall procedure of the data augmentation algorithm.}
\label{fig:augmentation} 
\end{figure*}

The data augmentation algorithm to synthesize turbulence-distorted video frames for training is summarized in Algorithm \ref{alg:distortion_blur_generation}. Figure \ref{fig:augmentation} illustrates the overall procedure of the data augmentation algorithm.

\subsection{{\bf WGAN}$-\ell_1$ with Multiframe Input}
The proposed turbulence removal network ({\bf TRN}) is a multi-frame subsampled {\bf WGAN} with the $\ell_1$ cost incorporated into the model. Multiframe input is adopted in {\bf TRN} to absorb sufficient information on the turbulence deformation of the original image. Then, {\bf TRN} is trained to remove geometric distortions and blurs with the {\bf WGAN} architecture. The additional $\ell_1$ cost attempts to retain the important textures of the original image.

\subsubsection{Multiframe Input}
We first discuss the input of our proposed {\bf TRN}. The conventional input for {\bf GAN}s is a noise vector $z \in \mathbb{R}^{N}$ randomly generated according to the normal distribution. Then the noise vector $z$ is transformed into the desired output through the generator. Our network is similar to DeBlurGan  \cite{kupyn2017deblurgan}. The architecture of DeBlurGan \cite{kupyn2017deblurgan} requires blurred image as an input and produce a deblurred image. Although blur is one of the consequences of turbulence observed in the frames, a single frame from a turbulence-distorted video as an input is experimentally shown to be ineffective in recovering the original image. Therefore, using the original architecture from DeBlurGan is insufficient to remove undesirable effects such as the geometric distortions. Motivated by this observation, the input in our network is a turbulence-distorted multiframe input originated from a clear image. Thus, the improved  version  of  our  new architecture  is  to  include  multiple frames  as  the  input. Instead of taking the whole video sequence as the input, subsampled frames are selected.

\begin{algorithm}[H]
\caption{Distortion and Blur Generation}
\label{alg:distortion_blur_generation} 
\begin{algorithmic}[1]
    \Statex Parameters: 
    \Statex $M = 1000 $ - number of iterations
    \Statex $N = 32 $ - patch size
    \Statex $\mu$ - mean of the Guassian kernel
    \Statex $\sigma$ - standard deviation of the Guassian kernel
    \Statex $S$ - distortion strength, uniform from [0.1,0.4]
    \Statex $B$ - blur constant, uniform from [0.1, 1]
    \Procedure{DistortBlur (Img, $\sigma$, $N$, $M$)}{}
    \State Create a Gaussian \text{kernel} from Normal CDF ($0.2 * \text{rand} - 1, \sigma$) 
    \For{$i = 1 \to M$}
        \State $ x \gets \text{randi} (\text{width} -2 *N) + N$
        \State $ y \gets \text{randi} (\text{height} -2 *N) + N$
        \State $u(x-N:x+N,y-N:y+N) $
        \Statex  $\hspace{2cm}\gets u(x-N:x+N,y-N:y+N) + \text{randn} * S$
        \State $v(x-N:x+N,y-N:y+N) $
        \Statex  $\hspace{2cm}\gets v(x-N:x+N,y-N:y+N) + \text{randn} * S$
        \State  Convolve the $u, v$ vector fields with the kernel
    \EndFor
    \State Wrap the image with $u, v$ vector fields with the \text{kernel}.
    \State Blur the image by convoluting it with a Gaussian smoothing window $w(n) = \exp(-\frac{n^2}{2 B^2})$ \\
    \Return Distorted Video Frames
    \EndProcedure
\end{algorithmic}
\end{algorithm}



With the data augmentation method described in last subsection, the training data is a multiple frames  $I_{TD} = (I_{TD}^{(1)}, I_{TD}^{(2)}, ..., I_{TD}^{(n)} )$ transformed from the original clean image $I_C$ of size $r \times s$. In {\bf TRN}, the input is a selected subsampled frames from $I_{TD}$.
Instead of using the whole sequence of frames as the input, we randomly select $ m = 20$ frames from the whole video in the training stage as the input for the generator in the GAN model. In the testing stage, we incorporate a subsampling method \cite{lau2017variational} to select the most useful frames as the input. The incorporating of the subsampling method into the network is shown to be effective in obtaining a significantly better restored image.

We now describe the subsampling method  we incorporate in the network in details. Given a turbulence-distorted video frames $ \mathcal{I}_{TD} = (I_{TD}^{(1)}, I_{TD}^{(2)}, ..., I_{TD}^{(n)}) $, we consider a variational model to get an optimal subsample set $\mathcal{J}$ of sharp and mildly distorted images. $J = \{ i_1, \cdots, i_m \}$ is the index set of the subsample set $\mathcal{J}$, where $m = \vert J \vert$ is the number of chosen video frames in the subsample. Simultaneously, we obtain a reference image $I_R$ from the subsample set $J$. The variational model is formulated in the following form:
\begin{equation} \label{eq: energy}
E(I_R,J) = \dfrac{1}{\vert J \vert} \bigg(\sum_{k \in J}\mathcal{F}( I_R, I_{TD}^{(k)}) + \lambda \mathcal{Q} (I_{TD}^{(k)}) \bigg)  - {\tau} \mathcal{R}(J)
\end{equation}
The fidelity term $\mathcal{F}$ is the discrepancy term between the reference image and the video frames. In our model, we define $\mathcal{F}( I_R, I_{TD}^{(k)}) = {\Big\Vert I_R - I_{TD}^{(k)} \Big\Vert}_2^2 $ for measuring the $\ell_2$ distance between the reference image $I_R$ and the subsampled video $\{I_{TD}^{(k)}\}_{k \in J}$. The quality term $\mathcal{Q}(I_{TD}^{(k)})$ for each video frame $I_{TD}^{(k)}$ is based on the normalized version of ${\Vert \Delta I_k \Vert}_1$:
\begin{equation}
\mathcal{Q}(I_{TD}^{(k)}) =  \dfrac{\underset{1\leq i \leq n}{\max} {\Vert \Delta I_{TD}^{(i)}  \Vert}_1 - {\Vert \Delta I_{TD}^{(k)} \Vert}_1}{\underset{1\leq i \leq n}{\max} {\Vert \Delta I_{TD}^{(i)}  \Vert}_1 - \underset{1\leq i \leq n}{\min} {\Vert \Delta I_{TD}^{(i)}  \Vert}_1 }
\end{equation}
The term $\Delta I_{TD}^{(k)}$ is the convolution of $I_{TD}^{(k)}$ with the Laplacian kernel specifically highlighting the edges and features of objects in the image $I_{TD}^{(k)}$. The sharper the image $I_{TD}^{(k)}$, the higher the magnitude of $\Delta I_{TD}^{(k)}$. As a consequence, the normalized quality measure $\mathcal{Q}(I_{TD}^{(k)})$ is smaller when $I_{TD}^{(k)}$ is sharp. The term $\lambda$ in the energy model $E(I_R,J)$ is a positive constant to quantify the importance of sharpness of the frame $I_{TD}^{(k)}$. The regularization term $\mathcal{R}$ is the concave increasing function $1 - e^{-\rho \vert J \vert}$, where $\rho > 0$ is a constant to quantify the importance of the number of selected frames. The function is chosen in order to acquire more information from additional video frames, whereas the effect on the quality of the reference image $I_R$ is reduced with a marginal increase in the subsample size. The detailed formulation of the variational model is described in \cite{lau2017variational}. An alternating minimization strategy can be used to solve the model, which is described in Algorithm \ref{alg:IS}.

\begin{algorithm}[H]
\caption{Image Subsampling}
\label{alg:IS} 
\begin{algorithmic}[1]
    \Statex Parameters: 
    \Statex $\lambda $ - sharpness parameter
    \Statex $\tau $ - subsample size parameter
    \Statex $\rho$ - subsample decay rate parameter
    \Procedure{Image Subsampling ($\mathcal{I}_{TD} = (I_{TD}^{(1)},I_{TD}^{(2)},\cdots, I_{TD}^{(n)}) $, $\lambda$, $\tau$, $\rho$)}{}
    \State Compute $I_{R}^0 = \dfrac{1}{n} \sum_{i=1}^n I_{TD}^{(i)}$
    \State Compute the quality measure $\mathcal{Q} (I_{TD}^{(k)})$ for each video frame $\{I_{TD}^{(k)} \}_{k=1}^n$
    \Repeat
       \State Given $J^{t-1}, I_{R}^{t-1}$. Fixing $I_{R}^{t-1}$, solve $$J^t = \underset{J}{\arg\min} \dfrac{1}{\vert J \vert} \bigg(\sum_{k \in J} {\Big\Vert I_R^t - I_{TD}^{(k)} \Big\Vert}_2^2  + \lambda \mathcal{Q} (I_{TD}^{(k)}) \bigg)  - {\tau} \bigg(1 - e^{-\rho \vert J \vert} \bigg)$$
        \State Compute $E_{1,k} = {\Big\Vert I_R^t - I_{TD}^{(k)} \Big\Vert}_2^2 +  \lambda \mathcal{Q} (I_{TD}^{(k)})$ for each $k$ and arrange $E_{1,k}$ in ascending order.
        \State Compute the sum $S_j$ for each $j$ and arrange $S_j$ in ascending order.
        \State $J^t \to \{k_1, k_2, \cdots, k_{j_1} \}$
        \State Fixing $J^t$, solve $$I^t = \underset{I}{\arg\min} \dfrac{1}{\vert J^t \vert} \bigg(\sum_{k \in J^t} {\Big\Vert I_R - I_{TD}^{(k)} \Big\Vert}_2^2 \bigg)$$
        \State $I_R^t \to \dfrac{1}{\vert J^t \vert} \sum_{k \in J^t} I_{TD}^{(k)}$
    \Until $E_1^{t-1} - E_1^{t} \leq \epsilon$\\
    \Return subsampled image sequence $\{ I_{TD}^{(k)} \}$
    \EndProcedure
\end{algorithmic}
\end{algorithm}
\subsubsection{U-Net Architecture for Generator Network}
The generator network $G$ we use is the U-Net \cite{ronneberger2015u}, which consists of five types of layers: convolutional layer, deconvolutional layer, max-pooling layer, Randomized Leaky ReLU activation layer ($\alpha = 0.2$) \cite{xu2015empirical} and instance normalization layer \cite{ulyanovinstance}. U-Net is known to involve a contracting path for contextual preservation and a symmetric expanding path for localization, and hence was particularly successful in image segmentation, denoising and super-resolution. Thus, U-net is used as the main architecture for our generator network $G$.\par
The subsampled turbulence-distorted multiframe passes through 7 blocks of convolutional layers and 6 blocks of deconvolutional layers to generate a clear image. The first 7 blocks $B_C^{(1)}, B_C^{(2)}, ...,  B_C^{(7)}$ contains convolutional layers, followed by the 6 remaining blocks $B_D^{(1)}, B_D^{(2)}, ..., B_D^{(6)}$ consisting of deconvolutional layers. Each block $B_C^{(i)}$ contains convolutional layers, non-linear activation layers and instance normalization layers. The temporal features extracted in each block are down-sampled by max-polling except for the features of the last block $B_C^{(7)}$. The features in $B_C^{(6)}$ and $B_C^{(7)}$ are concatenated before passing through the block $B_D^{(1)}$ in order to retain the deep features without too much information loss. The feature collected in the first block $B_D^{(1)}$ is then concatenated with the feature from the block $B_C^{(5)}$ to output the feature in the second block $B_D^{(2)}$. Repeating the process, we obtain a clear image $I_C$ which is of the same size as the original undistorted image. \par
The generator network is not pre-trained, since the input of the architecture is different from the conventional one. The conventional model takes the three channels of the image as input. In our case, we have a subsample of turbulence-distorted video frames $\mathcal{J}$, which are randomly chosen in the training stage and selected by the subsampling method introduced in the last subsection in the testing stage. The generator network $G$ is trained after the critic network $D$ is trained multiple times to give a clearer image $I_{TD} = G(I_{TD}^{(i_1)},...,I_{TD}^{(i_m)})$. The loss function of the generator network for removing geometric distortion and blurs is defined by 
\begin{equation}
L_G = - \sum_{n=1}^N  D(I_{TD}) +  \dfrac{\gamma}{N} {\Vert I_C - I_{TD} \Vert}_1
\end{equation}
The first term in the loss function $L_G$ is the adversarial loss that encourages solutions to reside on the manifold of natural images. In order to retain the textures inherited from the turbulence-distorted video frames, we further incorporate the $\ell_1$ loss into the loss function $L_G$. Note that the combination of the pixel-wise error term with the adversarial loss has an advantage. It was suggested that the minimization of the loss function that contains only the pixel-wise error term, such as the $\ell_1$ or $\ell_2$ error, is insufficient to produce a clear image \cite{ledig2017photo}. Besides, the $\ell_2$ error term can often cause image blur. Thus, we employ $\ell_1$ error, instead of the $\ell_2$ error, in our loss function $L_G$ to make the image much less blurry. Experimental results demonstrate that the combination of the two terms in the loss function $L_G$ can effectively remove geometric distortions and undesirable artifacts, such as image blurs.
\subsubsection{Critic Network}
The critic network $D$ in the {\bf WGAN} \cite{arjovsky2017wasserstein} is a deep {\bf CNN} involving convolutional layers, fully-connected layer, ReLU activation layer \cite{nair2010rectified} and instance normalization layer \cite{ulyanovinstance}. We denote the first 6 convolutional layers by $L^{(1)}, L^{(2)},...,L^{(6)}$ and the last fully connected layer by $L^{(7)}$. The critic values $D \circ G(I_{TD}))$ and $D \circ I_C$ are passed into the critic network to output the Wasserstain-1 distance
\begin{equation}
 \underset{ {\Vert D \Vert}_L \leq 1}{\max}  \quad \underset{x  \sim \mathbb{P}_r} { \mathbb{E} } [ D(x)] - \underset{\tilde{x}  \sim \mathbb{P}_g} { \mathbb{E} } [ D(\tilde{x})] 
\end{equation}
The critic network $D$ is trained till optimal before updating the generator network $G$. The loss function of the critic network $D$ in the training process is provided as follows:
\begin{equation}
L_D = D(G(I_{TD})) - D(G(I_C)) + \lambda {\bigg( {\Big\Vert \nabla D\Big(   \alpha I_C + (1-\alpha) I_{TD}  \Big) \Big\Vert}_2-1 \bigg)}^2,
\end{equation}
where $\alpha$ is a randomly generated number from uniform distribution $U[0,1]$. Since there is no pre-trained model involved in the generator network $G$, the whole training takes a long time and the loss blows up. In order to further enforce the $1$-Lipschitz assumption in the critic network $D$, we further impose the weight constraint $\lambda$. The weight is clipped in the interval $[-c, c]$. Together with this weight constraint, the training becomes more stable.

\section{Experiments}
\subsection{Dataset}
The dataset for training is collected from Flicker. It consists of 1500 images of buildings and 1000 images of chimneys. All collected images are resized to 256x256 and are synthetically deformed by our data augmentation algorithm. More specifically, each image is deformed to produce 100 deformed video sequences. Therefore, the whole dataset is enlarged by a factor of 100. We test the trained network on more than 400 testing data, which are different from the training dataset. The testing dataset consists of simulated video sequences as well as real turbulence-distorted video sequences. 

\subsection{Training Details}
\begin{figure*}[!htbp]
\centering
\includegraphics[scale=0.45]{{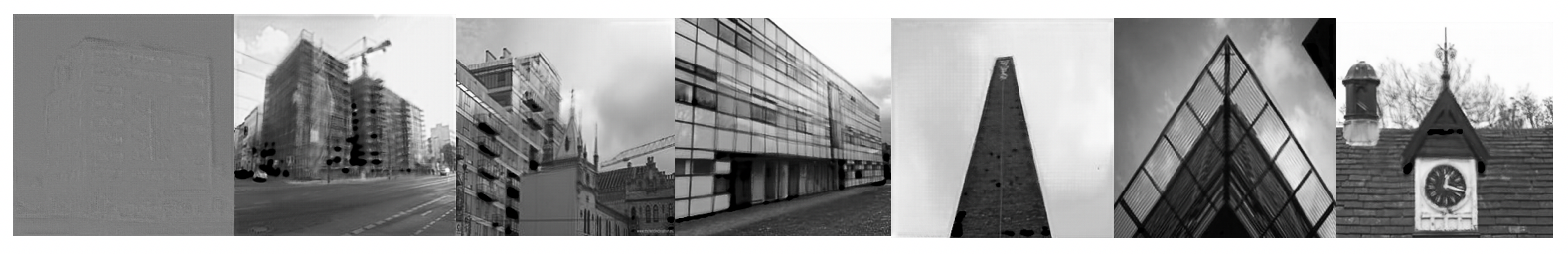}}
\caption{The training starting from the 1st epoch (left) to the 9th epoch (right). Each displayed image from its left image, except the first one, are generated for 1-2 epochs. The training performance is gradually improved.}
\label{fig:train}
\end{figure*}

The experiments are conducted in PyTorch \cite{paszke2017automatic} with a CUDA-enabled GPU. The data augmentation algorithm is carried out in Matlab before the deep learning process is conducted. The strength value of distortion $S$ and the blurring parameter $B$ are randomly sampled from $[0.1,0.4]$ and $[0.1,1]$ respectively. Adam solver \cite{kingma2014adam} is used for the gradient descent with a learning rate of $10^{-4}$, $\beta_1 = 0.5$ and  $\beta_2 = 0.99$ for both the generator $G_{\theta}$ and the critic $D_{\xi}$. We set 3 gradient descent steps for $D_{\xi}$ and then 1 step for $G_{\theta}$. We also apply the instance normalization and dropout to improve the training. In addition to the gradient penalty term \cite{gulrajani2017improved}, we enforce the parameters ${\xi}$ in the range $[-0.01,0.01]$. For each epoch, we train both the network with batch size of 1, and set $\lambda = 10, \gamma = 1000$. Furthermore, we randomly select 20 frames from the video sequence as our input. The whole training process for 40 epochs takes around 3 days. Figure \ref{fig:train} demonstrates that the restoration performance is gradually better in the training. The network firstly discards geometric distortion from the turbulence in the first few epochs and then attempts to deblur and preserve the texture of the original image in the remaining epochs. 

\section{Result}
After the {\bf TRN} is trained, we test its performance on more than 400 testing data consisting of simulated and real turbulence-distorted videos. The testing data are different from the training data. In this section, we report some of the experimental results.

\subsection{Restoration of simulated turbulence-distorted videos}
\begin{figure*}[!htbp]
\centering
\begin{tabular}{ccc}
\hspace{0.5cm} (a) Observed & \hspace{0.5cm} \ \ \ \ \ \  (b) {\bf TRN} (no sub)  &  \hspace{0.5cm}\ \ \ \ \  (c) {\bf TRN} (with sub) \\
\end{tabular}
\includegraphics[scale=0.45]{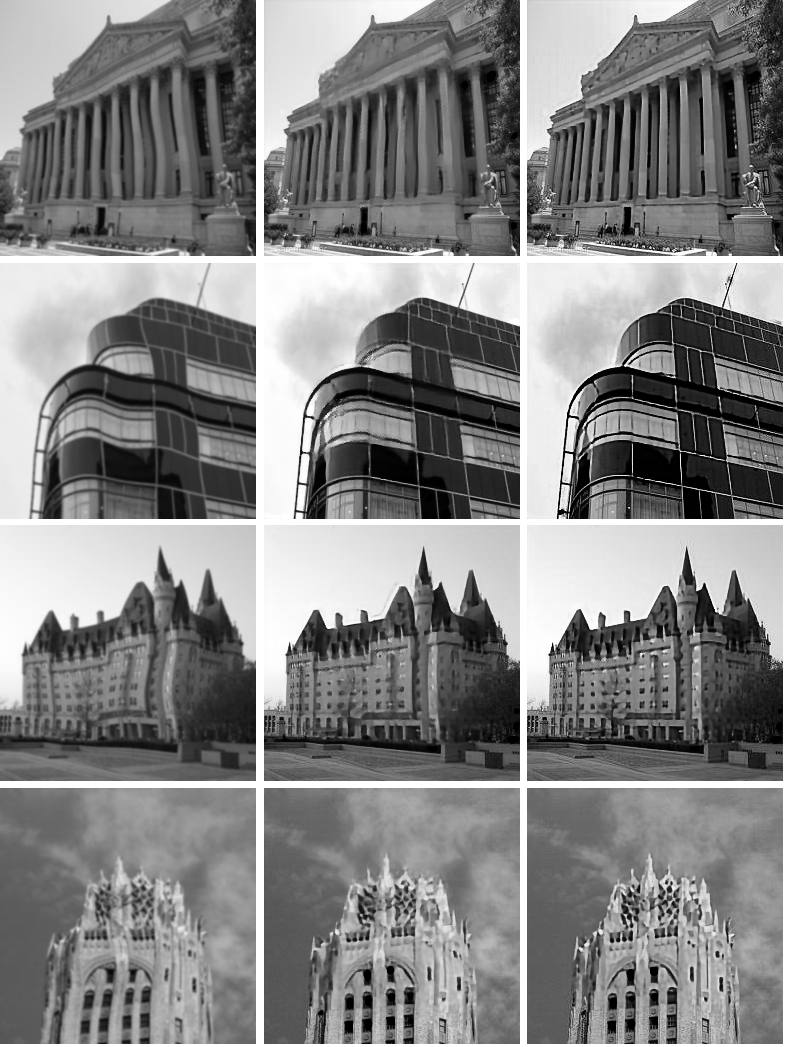}
\caption{Restoration of turbulence-distorted 'building' images. Column (a) shows the observed frames from each video. Column (b) shows the restoration results using the proposed {\bf TRN} without subsampling. Column (c) shows the restoration results using {\bf TRN} with subsampling.}
\label{fig:build_1} 
\end{figure*}

\begin{figure*}[!htbp]
\centering

\begin{tabular}{ccc}
\hspace{0.5cm} (a) Observed & \hspace{0.5cm} \ \ \ \ \ \  (b) {\bf TRN} (no sub)  &  \hspace{0.5cm}\ \ \ \ \  (c) {\bf TRN} (with sub) \\
\end{tabular}
\includegraphics[scale=0.45]{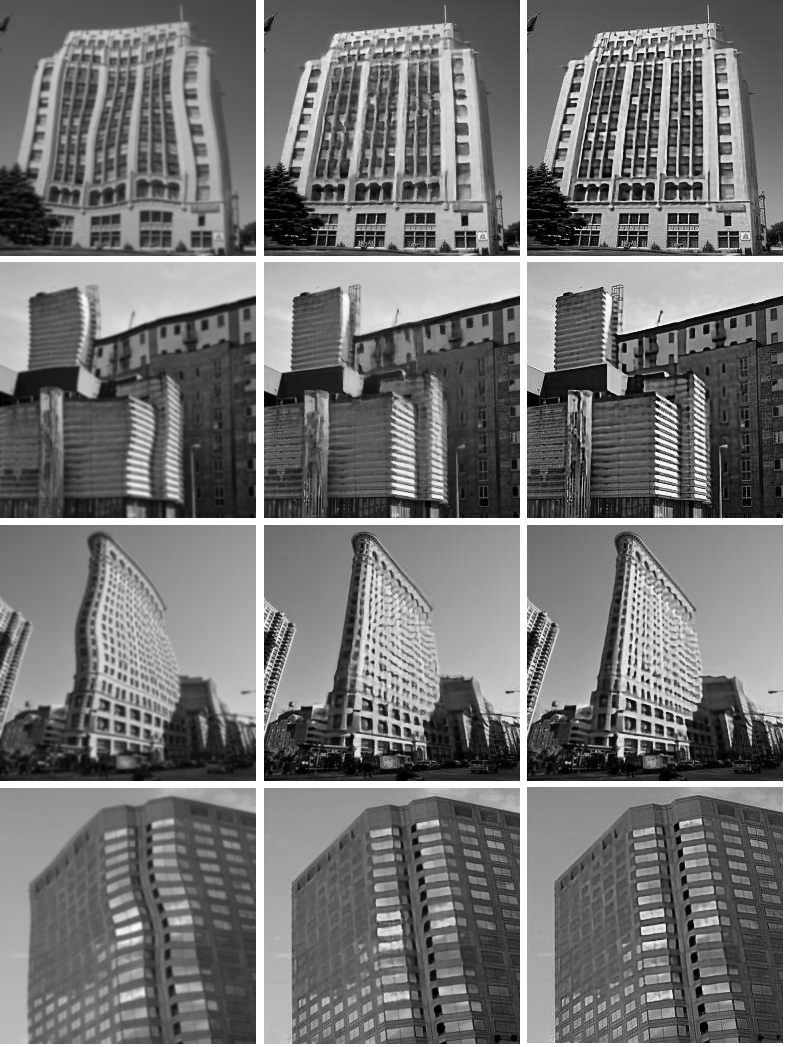}
\caption{Restoration of another set of turbulence-distorted 'building' images. Column (a) shows the observed frames from each video. Column (b) shows the restoration results using the proposed {\bf TRN} without subsampling. Column (c) shows the restoration results using {\bf TRN} with subsampling.}
\label{fig:build_2} 
\end{figure*}

Figure \ref{fig:build_1} and \ref{fig:build_2} show the restoration results of some simulated turbulence-distorted image sequences capturing different buildings. The first column shows the observed frames from each turbulence distorted image sequences, which are degraded by both geometric distortions and blurs. The middle column shows the restoration results using the {\bf TRN} without the incorporation of the subsampling method. Note that most geometric distortions and blurs are removed, although some amount of distortions can still be observed. The right column shows the restoration results using the {\bf TRN} with the incorporation of the subsampling method. With subsampling, the geometric distortions and blurs can be removed more successfully. The restoration results are more satisfactory compared to those without subsampling. It demonstrates the incorporation of the subsampling method into the deep network is beneficial.

\begin{figure*}[!htbp]
\centering

\begin{tabular}{ccc}
\hspace{0.5cm} (a) Observed & \hspace{0.5cm} \ \ \ \ \ \  (b) {\bf TRN} (no sub)  &  \hspace{0.5cm}\ \ \ \ \  (c) {\bf TRN} (with sub) \\
\end{tabular}
\includegraphics[scale=0.45]{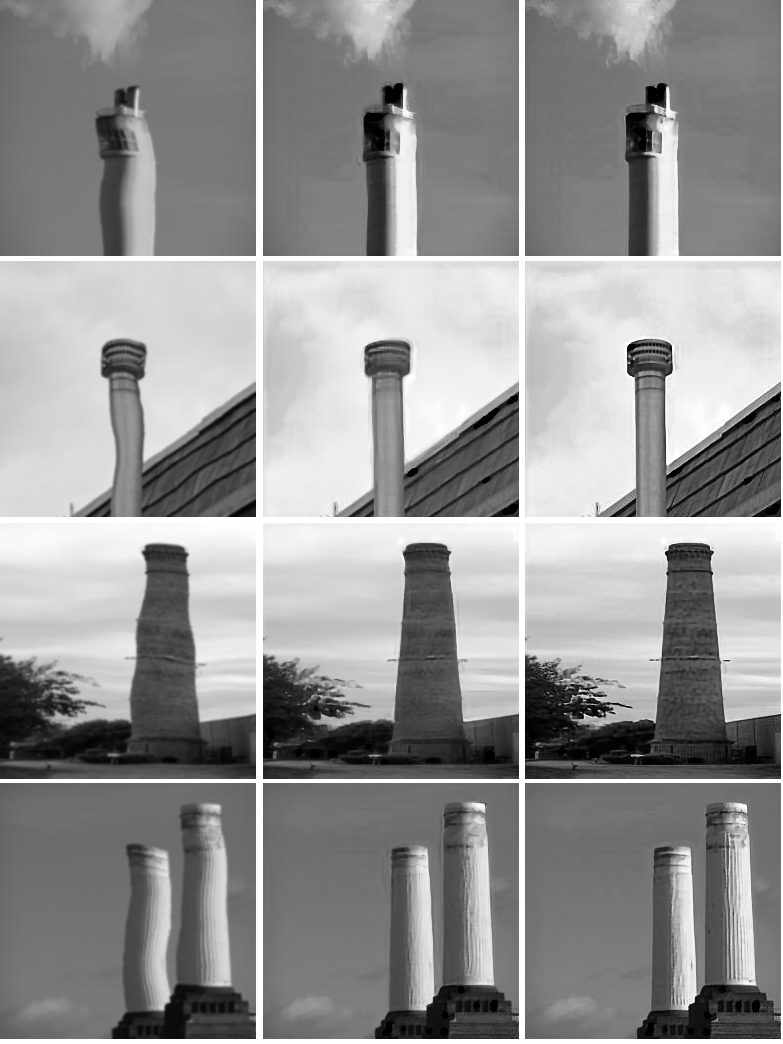}
\caption{Restoration of turbulence-distorted 'chimney' images. Column (a) shows the observed frames from each video. Column (b) shows the restoration results using the proposed {\bf TRN} without subsampling. Column (c) shows the restoration results using {\bf TRN} with subsampling.}
\label{fig:chimney_1} 
\end{figure*}

\begin{figure*}[!htbp]
\centering
\begin{tabular}{ccc}
\hspace{0.5cm} (a) Observed & \hspace{0.5cm} \ \ \ \ \ \  (b) {\bf TRN} (no sub)  &  \hspace{0.5cm}\ \ \ \ \  (c) {\bf TRN} (with sub) \\
\end{tabular}
\includegraphics[scale=0.45]{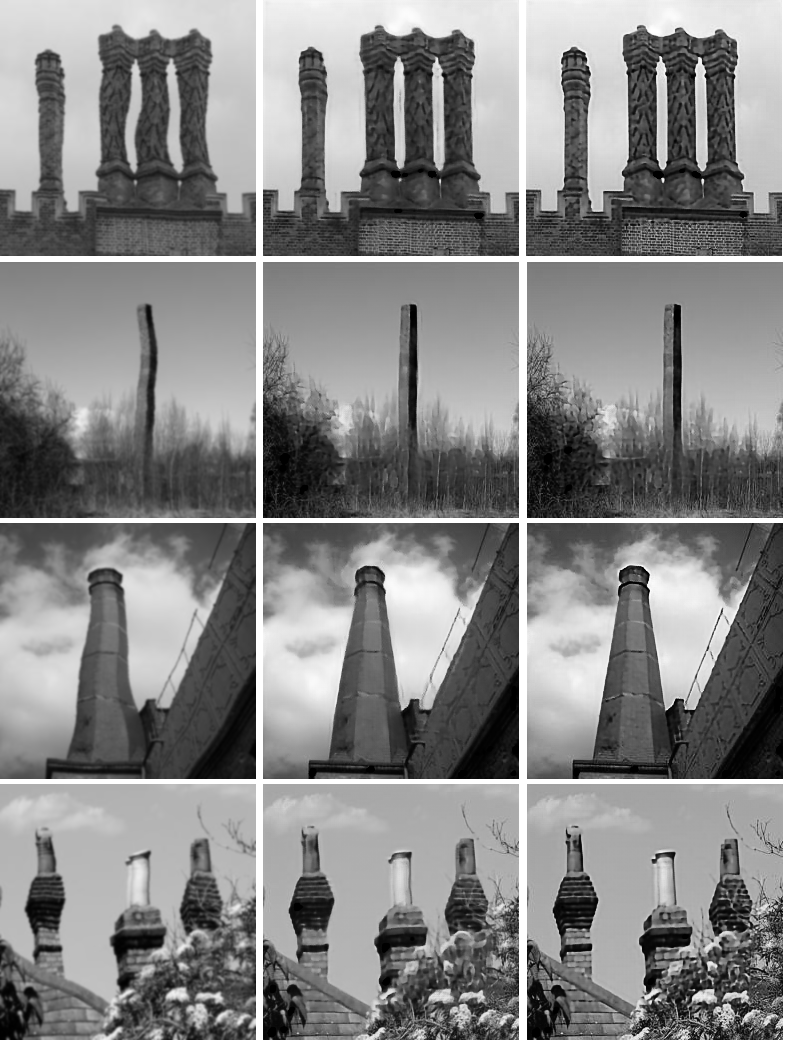}
\caption{Restoration of another set of turbulence-distorted 'chimney' images. Column (a) shows the observed frames from each video. Column (b) shows the restoration results using the proposed {\bf TRN} without subsampling. Column (c) shows the restoration results using {\bf TRN} with subsampling.}
\label{fig:chimney_2} 
\end{figure*}

We have test our deep network on some 'chimney' image sequences. Figure \ref{fig:chimney_1} and \ref{fig:chimney_2} shows the restoration results of some simulated turbulence-distorted image sequences capturing different chimneys. Again, the first column shows the observed frames from each turbulence distorted image sequences, which are degraded by both geometric distortions and blurs. The middle column shows the restoration results using the {\bf TRN} without the incorporation of the subsampling method. The right column shows the restoration results using the {\bf TRN} with the incorporation of the subsampling method. Again, with the incorporation of the subsampling method, the geometric distortions and blurs can be removed more successfully. The restoration results are more satisfactory compared to those without subsampling. It again demonstrates the benefit of incorporating the subsampling method into the deep network.

\begin{figure*}[!htbp]
\centering

\begin{tabular}{cc}
 (a) Observed & \ \ \ \ \ (b) Restored by {\bf TRN}  \\
\end{tabular}
\includegraphics[scale=0.45]{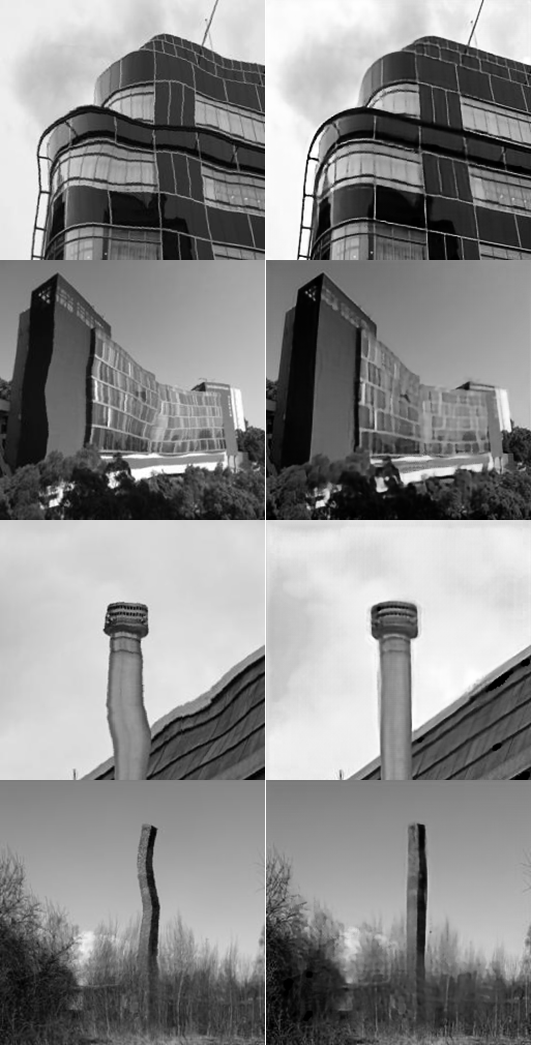}
\caption{Restoration of image sequence distorted by large quasi-conformal deformations. Column (a) shows the observed frames from each video. Column (b) shows the restoration results using the proposed {\bf TRN}.}
\label{fig:mu_build_1} 
\end{figure*}

To test the performance of our trained network to handle general large deformations, we randomly generate geometric distortions of an original image using large quasi-conformal deformations. More specifically, we randomly select some pixel positions in the image domain. A patch-wise triangular mesh is formed with the chosen position as the center. The method we use to generate artificial turbulence is deformation using Laplace-Beltrami solver (LBS) \cite{lam2014landmark}. We propose to assign the Beltrami coefficient $\mu$ which is a measure of nonconformality on each face vertex as follows:
$$\mu = \bigg[(0.6 + \epsilon_1) \cos \Big((4 + \epsilon_2)\pi x^2 \Big) \bigg] + \bigg[(0.6 + \epsilon_3) \sin \Big((6 + \epsilon_4)\pi y^2\Big) \cos \Big((8 + \epsilon_5)\pi x y\Big) \bigg] i,$$
where $\epsilon_i$ are numbers randomly chosen in the range $[0,0.3]$ for $i=1,2$ and $[-1,1]$ for $i=3,4,5$. Then we obtain the deformation field by using the LBS solver, and wrap the image. By introducing image blurs to each deformed images, we obtain an image sequence with large geometric distortions and blurs. Note that the quasi-conformal deformations have never been seen in the training process. Our aim is to investigate whether the trained deep network can deal with general deformations. The experimental results are shown in Figure \ref{fig:mu_build_1}. In the Figure \ref{fig:mu_build_1}, the first column shows the observed frames from each distorted image sequences with large quasi-conformal deformations. The second column shows the restored images using {\bf TRN}. The geometric distortions and blurs are successfully removed. These results show that the {\bf TRN} can effectively handle general large deformations.

\subsection{Comparison with other methods}
\begin{figure*}[!htbp]
\centering
\begin{tabular}{ccc}
 (a) Restored by {\bf TRN} & \hspace{5mm} (b) Restored by {\bf SGL} & \hspace{3mm} (c) Restored by {\bf IRIS} \\
\end{tabular}
\includegraphics[scale=0.45]{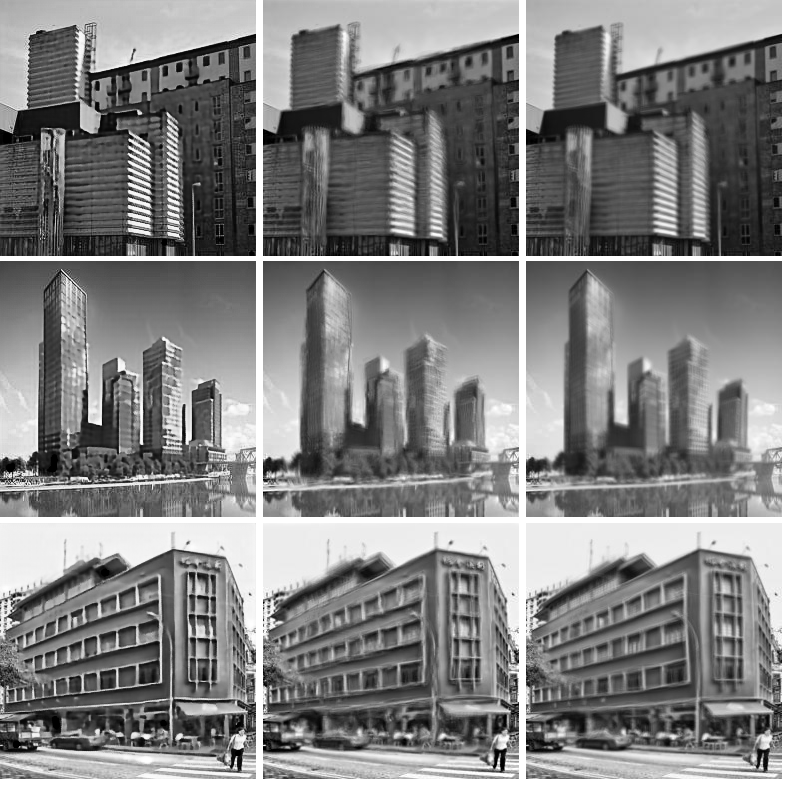}
\caption{Comparison between {\bf TRN}, {\bf SGL} and {\bf IRIS} on 'building 1', 'building 2' and 'building 3'. (a) shows the restoration results by {\bf TRN}. (b) shows the restoration results by {\bf SGL}. (b) shows the restoration results by {\bf IRIS}}. 
\label{fig:comparison} 
\end{figure*}

We also compare our proposed deep-learning based algorithms with other existing methods, namely, the {\bf SGL} method \cite{lou2013video} and the {\bf IRIS} method \cite{lau2017variational}. Some experimental results are shown in Figure \ref{fig:comparison}. In the Figure \ref{fig:comparison}, the first column shows the restoration results of some turbulence-distorted image sequences using {\bf TRN}. The second column shows the restoration results using {\bf SGL}. The last column shows the restoration results using {\bf IRIS}. The restoration results using {\bf SGL} is generally blurry and geometrically distorted. The results restored by {\bf IRIS} have less geometric distortions and blurs, although some geometric deformations can still be visualized. In general, {\bf TRN} gives the best restoration results with least geometric distortions and blurs. These visual results are also validated quantitatively using PSNR and SSIM, as reported in Table \ref{tab: comp_table}.  

\begin{table*}[!htbp]
\centering
\begin{tabular}{ccccccccc}
   \toprule
   & & PSNR & & &  & SSIM &\\
   \midrule
    & {\bf TRN} & {\bf SGL} & {\bf IRIS} & & {\bf TRN} & {\bf SGL} & {\bf IRIS} \\
   \midrule
   building 1  & 21.3 & 19.3 & 20.1 &    & 0.838 & 0.697 & 0.749  \\  
     building 2  & 23.7 & 22.1 & 23.0 &   & 0.808 & 0.714 & 0.740 \\
     building 3  & 24.7 & 23.8 & 24.5 &   & 0.836 & 0.770 & 0.793 \\
\end{tabular} 
\caption{PSNR and SSIM of the restored images using different deturbulence models.}
\label{tab: comp_table}
\end{table*}

\subsection{Restoration of real turbulence-distorted videos}
\begin{figure*}[!htbp]
\centering
\begin{tabular}{ccc}
(a) Observed  &\ \ \ \ \  \ \ \ \ (b) {\bf TRN} (no sub) & \ \ \ \ \ (c) {\bf TRN} (with sub)
\end{tabular}
\includegraphics[scale=0.4]{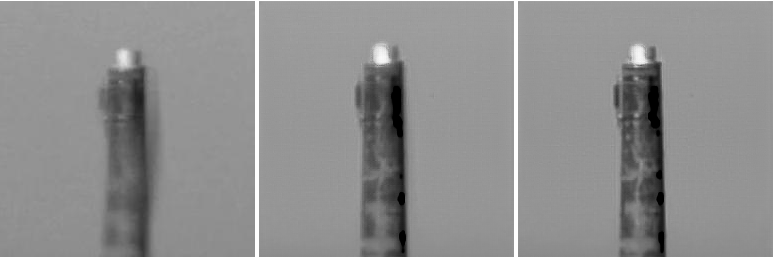}
\caption{Restoration of real turbulence-distorted image sequence capturing a chimney. (a) shows an observed frame from the image sequence. (b) shows the restored image using {\bf TRN} without subsampling. (c) shows the restored image using {\bf TRN} with subsampling.}
\label{fig:real_chimney} 
\end{figure*}

We also test the {\bf TRN} on real turbulence-distorted videos that do not have a clear ground-truth image. Figure \ref{fig:real_chimney} shows the restoration results of a real `chimney' turbulence-distorted image sequence. (a) shows an observed frame from the image sequence. (b) shows the restoration results using {\bf TRN} without subsampling. Most geometric distortions and blurs are suppressed. (c) shows the restoration results using {\bf TRN} with subsampling. With subsampling, the results are more satisfactory compared to those without subsampling.  It again demonstrates the effectiveness of incorporating the subsampling model into the deep network.

\begin{figure*}[!htbp]
\centering
\begin{tabular}{ccc}
(a) Observed  \ \ \ \ \ \ \ \ \ \  (b) {\bf TRN} (no sub) & \ \ \ \ \ (c) {\bf TRN} (with sub)\\
\end{tabular}
\includegraphics[scale=0.4]{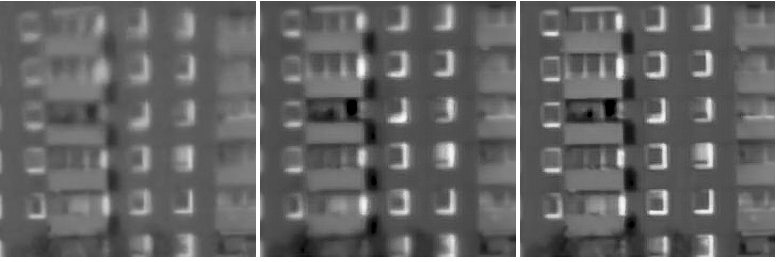}
\caption{Restoration of real turbulence-distorted image sequence capturing a building. (a) shows an observed frame from the image sequence. (b) shows the restored image using {\bf TRN} without subsampling. (c) shows the restored image using {\bf TRN} with subsampling.}
\label{fig:real_building} 
\end{figure*}

Figure \ref{fig:real_building} shows the restoration results of another real turbulence-distorted image sequence capturing a building. Again, (a) shows an observed frame from the image sequence. (b) shows the restoration results using {\bf TRN} without subsampling. (c) shows the restoration results using {\bf TRN} with subsampling. As before, with subsampling, the results are more satisfactory than those without subsampling.

\section{Conclusion}
We introduce the turbulence removal network ({\bf TRN}), which is a Generative Adversarial Network ({\bf GAN}) incorporated with $\ell_1$ objective function, to suppress geometric distortions as well as removing blurs of image sequences distorted by turbulence. Although there is only a limited amount of available data corrupted by real turbulence, we proposed a data augmentation method to synthetically generate turbulence-distorted image frames for training. A subsampling method is further incorporated into the trained network to obtain an improved restoration result. Extensive experiments have been carried out to test the deep network, which demonstrates the effectiveness of the proposed model to restore turbulence-distorted images. In the future, we will explore the possibility to develop a turbulence removal network to restore turbulence-distorted video with moving objects.

\section*{Acknowledgment}
We would like to thank Mr. M. Hirsch and Dr. S. Harmeling from Max Planck Institute for Biological Cybernetics for sharing the real chimney and building video sequence. Lok Ming Lui is supported by HKRGC GRF (Project ID: 402413).   

\section*{References}

\bibliography{mybibfile}




\end{document}